\documentclass{article}
\usepackage[latin9]{inputenc}
\usepackage{geometry}
\usepackage{amsmath}
\usepackage{amssymb}
\usepackage{graphicx}

\makeatletter

\providecommand{\tabularnewline}{\\}

\@ifundefined{date}{}{\date{}}
\makeatother

\begin{document}

\title{Boosted Markov Networks for Activity Recognition}

\author{{\small $^{\#}$~}Tran The Truyen{\small $~^{1}$}, Hung Hai Bui{\small $~^{2}$},
Svetha Venkatesh{\small $~^{3}$}\\
 \textit{Department of Computing, Curtin University of Technology
}\\
\textit{ GPO Box U 1987, Perth, WA, Australia. }\\
\textit{ }\textit{\small $^{1,3}$}\textit{~\{trantt2,svetha\}@cs.curtin.edu.au,
}\textit{\small $^{2}$}\textit{bui@ai.sri.com }}
\maketitle
\begin{abstract}
We explore a framework called boosted Markov networks to combine the
learning capacity of boosting and the rich modeling semantics of Markov
networks and applying the framework for video-based activity recognition.
Importantly, we extend the framework to incorporate hidden variables.
We show how the framework can be applied for both model learning and
feature selection. We demonstrate that boosted Markov networks with
hidden variables perform comparably with the standard maximum likelihood
estimation. However, our framework is able to learn sparse models,
and therefore can provide computational savings when the learned models
are used for classification. 
\end{abstract}

\section{Introduction}

Recognising human activities using sensors is currently a major challenge
in research. Typically, the information extracted directly from sensors
is either not discriminative enough or is too noisy to infer activities
occurring in the scene. Human activities are complex and evolve dynamically
over time. Temporal probabilistic models such as hidden Markov models
(HMMs) and dynamic Bayesian networks (DBNs) have been the dominant
models used to solve the problem \cite{Aggarwal-Cai-CVIU99,Bui-et-alJAIR02,Yamato-et-alCVPR92}.
However, these models make a strong assumption in the generative process
by which the data is generated in the model. This makes the representation
of complex sensor data very difficult, and possibly results in large
models.

Markov networks (MNs) (also known as Markov random fields) offer an
alternative approach, especially in form of conditional random fields
(CRFs) \cite{lafferty01conditional}. In CRFs, the observation is
not modelled, and so we have the freedom to incorporate overlapping
features, multiple sensor fusion, and long-range dependencies into
the model. The discriminative nature and the underlying graphical
structure of the MNs make it especially suitable to the problem of
human activity recognition.

Boosting is a general framework to gradually improve the performance
of the weak learner (which can be just slightly better than a random
guess). A popular version called AdaBoost \cite{Fre_Sch96,Sch98,schapire99improved}
forces the weak learner to focus more on hard-to-learn examples from
the examples seen so far. The final strong learner is the weighted
sum of all weak learners added to this ensemble in each iteration.
Most of the work so far with boosting involves only unstructured output,
except for a few occasions, such as the work in \cite{AA63,Dietterich-et-alICML04,NIPS2005_704}.

We are motivated to use boosting for parameter estimation of Markov
networks (which we call boosted Markov networks (BMNs)), as recent
results have shown the close relationship between boosting and the
maximum likelihood estimation (MLE) \cite{Friedman-et-al00,nips02-LT09}.
Furthermore, we use the inherent capacity of boosting for feature
selection integrated with learning. We are motivated by studies that
the typical log-linear models imposed on Markov networks can easily
be overfitted with little data and many irrelevant features, but can
be overcome by the use of explicit feature selection, either independently
or integrated with learning (e.g. see \cite{Pietra-inducing-97}).

Previous work \cite{AA63,Dietterich-et-alICML04,NIPS2005_704} have
explored the BMNs showing promising results. The BMNs integrate the
discriminative learning power of boosting and the rich semantics of
the graphical model of Markov networks. In this paper, we further
explore alternative methods for this approach. To handle hidden variables
as in the standard MLE, we extend the work by Altun \textit{et al.}
\cite{AA63}. This is a variant of the multiclass boosting algorithm
called AdaBoost.MR \cite{Fre_Sch96,schapire99improved}. We also suggest
an approximation procedure for the case of intractable output structures.
The proposed framework is demonstrated through our experiments in
which boosting can provide a comparable performance to MLE. However,
since our framework uses sparse features it has the potential to provide
computational savings during recognition.

The novelty of the paper is two-fold: a) We present the first work
in applying boosting for activity recognition, and b) we derive a
boosting procedure for structured models with missing variables and
use a parameter update based on quadratic approximation instead of
loose upper bounds as in \cite{AA63} to speed up the learning.

The organisation of the paper is as follows. The next section reviews
related work. Section \ref{sec:BMN} presents the concept of boosted
Markov networks (BMNs) detailing how boosting is employed to learn
parameters of Markov networks. Section \ref{sec:comp} goes into the
details of efficient computation for tree-structured and general networks.
We then describe our experiments and results in applying the model
to video-based human activity recognition in an indoor environment.
The last section discusses any remaining issues.

%--------

\section{Related work}

Our work is closely related to that in \cite{Dietterich-et-alICML04},
where boosting is applied to learn parameters of the CRFs using gradient
trees \cite{Friedman-ann-stat01}. The objective function is the log-likelihood
in the standard MLE setting, but the training is based on fitting
regression trees in a stage-wise fashion. The final decision function
is in the form of a linear combination of regression trees. \cite{Dietterich-et-alICML04}
employs functional gradients for the log-loss in a manner similar
to LogitBoost \cite{Friedman-et-al00}, whilst we use the original
gradients of the exponential loss of AdaBoost \cite{Fre_Sch96,Sch98,schapire99improved}.
Thus \cite{Dietterich-et-alICML04} learns the model by maximising
the likelihood of the data, while we are motivated by minimising the
errors in the data directly. Moreover, \cite{Dietterich-et-alICML04}
indirectly solves the structured model learning problem of MLE by
converting the problem to a standard unstructured learning problem
with regression trees. In contrast, we solve the original structured
learning problem directly without the structured-unstructured conversion.
In addition, the paper in \cite{Dietterich-et-alICML04} does not
incorporate hidden variables as our work does.

Another work, \cite{NIPS2005_704}, integrates the message passing
algorithm of belief propagation (BP) with a variant of LogitBoost
\cite{Friedman-et-al00}. Instead of using the per-network loss as
in \cite{Dietterich-et-alICML04}, the authors of \cite{NIPS2005_704}
employ the per-label loss (e.g. see \cite{Altun-03} for details of
the two losses), that is they use the marginal probabilities. Similar
to \cite{Dietterich-et-alICML04}, \cite{NIPS2005_704} also converts
the structured learning problem into a more conventional unstructured
learning problem. The algorithm thus alternates between a message
passing round to update the local per-label log-losses, and a boosting
round to update the parameters. However, as the BP is integrated in
the algorithm, it is not made clear on how to apply different inference
techniques when the BP fails to converge in general networks. It is
also unclear on how to extend the method to deal with hidden variables.

There have been a number of attempts to exploit the learning power
of boosting applied to structured models other than Markov networks,
such as dynamic Bayesian networks (DBNs) \cite{Garg-et-al-IEEE03},
Bayesian network classifier \cite{Jing-et-alICML05}, and HMMs \cite{Yin-et-al-CVPR04}.
%The authors report
%improvement over the original model without boosting, thus motivating
%us to further investigate the boosted Markov networks.

%-----------

\section{Boosted Markov networks}

\label{sec:BMN}

\subsection{Markov networks}

We are interested in learning a structured model in which inference
procedures can be performed. A typical inference task is decoding,
e.g. to find the most probable label set (configuration) $y*$ for
the network given an observation $x$ $y*=\arg\max_{y}p(y|x)$. For
a single node network, this is often called the classification problem.
For a network of $N$ nodes, the number of configurations is exponentially
large. We assume the conditional exponential distribution (a.k.a.
log-linear model) 
\begin{eqnarray}
p(y|x)=\frac{1}{Z(x)}\exp(F(x,y))\label{exp-model}
\end{eqnarray}
where $Z(x)=\sum_{y}\exp(F(x,y))$ is the normalisation factor, and
$F(x,y)=\sum_{k}\lambda_{k}f_{k}(x,y)$. $\{f_{k}\}$ are features
or weak hypotheses in boosting literature. This conditional model
is known as the conditional random fields (CRFs) \cite{lafferty01conditional}.
The decoding reduces to $y*=\arg\max_{y}F(x,y)$.

Often, in Markov networks the following decomposition is used 
\begin{eqnarray}
f_{k}(x,y)=\sum_{c}f_{k}(x,y_{c})\label{additive-form}
\end{eqnarray}
where $c$ is the index of the clique defined by the structure of
the network. This decomposition is essential to obtain a factorised
distribution, which is vital for efficient inference in tree-like
structures using dynamic programming.

Denote by $y=(v,h)$ the visible and hidden variables (which are represented
by nodes in the network). Given $n$ i.i.d observations $\{x_{i}\}_{i=1}^{n}$,
maximum likelihood learning in MNs minimises the log-loss 
\begin{eqnarray}
L_{log}=-\sum_{i}\log p(v^{i}|x^{i})=\sum_{i}\log\frac{1}{p(v^{i}|x^{i})}\label{log-loss}
\end{eqnarray}

\subsection{Exponential loss for incomplete data}

We view the activity labeling and segmentation problems as classification
where the number of distinct classes is exponentially large, i.e.
$|Y|^{N}$, where $|Y|$ is the size of label set, and $N$ is the
number of nodes in the Markov network. Following the development by
Altun et al. \cite{AA63}, we define a new \textit{expected ranking
loss} \cite{schapire99improved} to incorporate hidden variables as
follows 
\begin{eqnarray}
L_{rank}=\sum_{i}\sum_{h}p(h|v^{i},x^{i})\sum_{v\neq v^{i}}\delta[\Delta F(x^{i},v,h)>0]\label{rank-loss}
\end{eqnarray}
where $\Delta F(x^{i},v,h)=F(x^{i},v,h)-F(x^{i},v^{i},h)$, and $\delta[z]$
is the indicator function of whether the predicate $z$ is true. This
rank loss captures the expectation of moments in which the classification
is wrong because if it is true, then $\max_{v}F(x^{i},v,h)=F(x^{i},v^{i},h)$
implying $F(x^{i},v,h)<F(x^{i},v^{i},h)\forall v\ne v^{i}$.

However, it is much more convenient to deal with a smooth convex loss
and thus we formulate an upper bound of the rank loss, e.g. the exponential
loss 
\begin{eqnarray}
L_{exp}=\sum_{i}\sum_{h}p(h|v^{i},x^{i})\sum_{v}\exp(\Delta F(x^{i},v,h))\label{exp-loss}
\end{eqnarray}
It is straightforward to check that (\ref{exp-loss}) is indeed an
upper bound of (\ref{rank-loss}). It can be seen that (\ref{rank-loss})
includes the loss proposed in \cite{AA63} as a special case when
all variables are observed, i.e. $y=v$. It is essentially an adapted
version of AdaBoost.MR proposed in \cite{schapire99improved}.

A difficulty with this formulation is that we do not know the true
conditional distribution $p(h|v^{i},x^{i})$. First, we approximate
it by the learned distribution at the previous iteration. Thus, the
conditional distribution is updated along the way, starting from some
guessed distribution, for example, a uniform distribution. Second,
we assume the log-linear model as in (\ref{exp-model}), leading to
\begin{eqnarray*}
\sum_{v}\exp(\Delta F(x^{i},v,h))=\frac{\sum_{v}\exp(F(x^{i},v,h))}{\exp(F(x^{i},v^{i},h))}=\frac{1}{p(v^{i}|h,x^{i})}
\end{eqnarray*}
which can be fed into (\ref{exp-loss}) to obtain 
\begin{eqnarray}
L_{exp}=\sum_{i}\sum_{h}\frac{p(h|v^{i},x^{i})}{p(v^{i}|h,x^{i})}=\sum_{i}\frac{1}{p(v^{i}|x^{i})}\label{exp-loss2}
\end{eqnarray}
We can notice the similarity between the exponential loss in (\ref{exp-loss2})
and the log-loss in (\ref{log-loss}) as log(.) is a monotonically
increasing function. The difference is the exponential scale used
in (\ref{exp-loss2}) with respect to features $\{f_{k}\}$ compared
to the linear scale in (\ref{log-loss}).

\subsection{Boosting-based learning}

The typical boosting process has many rounds, each of which selects
the best weak hypothesis and finds the weight for this hypothesis
to minimise the loss. Translated in our context, the boosting-based
learning searches for the best feature $f_{j}$ and its coefficient
to add to the ensemble $F^{t+1}=F^{t}+\alpha^{t}f_{j}$ so that the
loss in (\ref{exp-loss}) is minimised. 
\begin{eqnarray}
(\alpha^{t},j) & = & \arg\min_{\alpha,k}L_{exp}(t,\alpha,k),\mbox{ where}\label{exp-min}\\
L_{exp}(t,\alpha,k) & = & \sum_{i}E_{h|v^{i},x^{i},t}[\sum_{v}\exp(\Delta F^{i,t}+\alpha\Delta f_{k}^{i})]\nonumber 
\end{eqnarray}
where $E_{h|v^{i},x^{i},t}[z(h)]=\sum_{h}p(h|v^{i},x^{i},t)z(h)$;
$F^{i,t}$ and $f_{k}^{i}$ are shorthands for $F^{t}(x^{i},v,h)$
and $f_{k}(x^{i},v,h)$, respectively. Note that this is just an approximation
to (\ref{exp-loss}) because we fix the conditional distribution $p(h|v^{i},x^{i},t)$
obtained from the previous iteration. However, this still makes sense
since the learning is incremental, and thus the estimated distribution
will get closer to the true distribution along the way. Indeed, this
captures the essence of boosting: during each round, the weak learner
selects the weak hypothesis that best minimises the following loss
over the weighted data distribution (see \cite{schapire99improved})
\begin{eqnarray}
(\alpha^{t},j)=\arg\min_{\alpha,k}\sum_{i}\sum_{v,h}D(i,v,h,t)\exp(\alpha\Delta f_{k}^{i})\label{weighted-loss}
\end{eqnarray}
where the weighted data distribution $D(i,v,h,t)=p(h|v^{i},x^{i},t)\exp(\Delta F^{i,t})/Z(i,t)$
with $Z(i,t)$ being the normalising constant. Since the data distribution
does not contain $\alpha$, (\ref{weighted-loss}) is identical to
(\ref{exp-min}) up to a constant.

\subsection{Beam search}

\label{sec:beam} It should be noted that boosting is a very generic
framework to boost the performance of the weak learner. Thus we can
build more complex and stronger weak learners by using some ensemble
of features and then later fit them into the boosting framework. However,
here we stick to simple weak learners, which are features, to make
the algorithm compatible with the MLE.

We can select a number of top features and associated coefficients
that minimise the loss in (\ref{weighted-loss}) instead of just one
feature. This is essentially a beam search with specified beam size
$S$.

\subsection{Regularisation}

%It was commonly believed that boosting resists overfitting.
%However, it has been recently shown in more extensive simulations
%that boosting can overfit, if little and noisy training data is fed to
%the algorithm. In the human activity domain, this is often the case
%because manual labeling is expensive, and there are usually no clear
%cut boundaries between successive activities. The activities can even
%be partially overlapping. Thus no perfect labeling can be given, and
%regularisation is therefore needed
%in the boosting framework.

We employ the $l_{2}$ regularisation term to make it consistent with
the popular Gaussian prior used in conjunction with the MLE of Markov
networks. It also maintains the convexity of the original loss. The
regularised loss becomes 
\begin{eqnarray}
L_{reg}=L_{non-reg}+\sum_{k}\frac{\lambda_{k}^{2}}{2\sigma_{k}^{2}}
\end{eqnarray}
where $L_{non-reg}$ is either $L_{log}$ for MLE in (\ref{log-loss})
or $L_{exp}$ for boosting in (\ref{exp-loss}). Note that the regularisation
term for boosting does not have the Bayesian interpretation as in
the MLE setting but is simply a constraint to prevent the parameters
from growing too large, i.e. the model fits the training data too
well, which is clearly sub-optimal for noisy and unrepresentative
data. The effect of regularisation can be numerically very different
for the two losses, so we cannot expect the same $\sigma$ for both
MLE and boosting.

%----------

\section{Efficient computation}

\label{sec:comp} Straightforward implementation of the optimisation
in (\ref{exp-min}) or (\ref{weighted-loss}) by sequentially and
iteratively searching for the best features and parameters can be
impractical if the number of features is large. This is partly because
the objective function, although it can be tractable to compute using
dynamic programming in tree-like structures, is still expensive. We
propose an efficient approximation which requires only a few vectors
and an one-step evaluation. The idea is to exploit the convexity of
the loss function by approximating it with a convex quadratic function
using second-order Taylor's expansion. The change due to the update
is approximated as 
\begin{eqnarray}
\Delta J(\alpha,k)\approx\left.\frac{dJ(\alpha,k)}{d\alpha}\right|_{\alpha=0}\alpha+\left.\frac{1}{2}\frac{d^{2}J(\alpha,k)}{d\alpha^{2}}\right|_{\alpha=0}\alpha^{2}\label{Taylor-approx}
\end{eqnarray}
where $J(\alpha,k)$ is a shorthand for $L(t,\alpha,k)$. The selection
procedure becomes $(\alpha^{t},j)=\arg\min_{\alpha,k}J(\alpha,k)=\arg\min_{\alpha,k}\Delta J(\alpha,k)$.
The optimisation over $\alpha$ has an analytical solution $\alpha_{t}=-J'/J''$.

Once the feature has been selected, the algorithm can proceed by applying
an additional line search step to find the best coefficient as $\alpha^{t}=\arg\min_{\alpha}L(t,\alpha,j)$.
One way to do is to repeatedly apply the update based on (\ref{Taylor-approx})
until convergence.

Up to now, we have made an implicit assumption that all computation
can be carried out efficiently. However, this is not the case for
general Markov networks because most quantities of interest involve
summation over an exponentially large number of network configurations.
Similar to \cite{AA63}, we show that dynamic programming exists for
tree-structured networks. However, for general structures, approximate
inference must be used.

There are three quantities we need to compute: the distribution $p(v_{i}|x_{i})$
in (\ref{exp-loss2}), the first and second derivative of $J$ in
(\ref{Taylor-approx}). For the distribution, we have $p(v^{i}|x^{i})=\sum_{h}p(v^{i},h|x^{i})=Z(v^{i},i)/Z(i)$
where $Z(v^{i},i)=\sum_{h}\exp(\sum_{c}F(x^{i},v_{c}^{i},h_{c}))$
and $Z(i)=\sum_{y}\exp(\sum_{c}F(x^{i},y_{c}))$. Both of these partition
functions are in the form of sum-product, thus, they can be computed
efficiently using a single pass through the tree-like structure. The
first and second derivatives of $J$ are then 
\begin{eqnarray}
J'|_{\alpha=0} & = & \sum_{i}E_{h|v^{i},x^{i},t}[\sum_{v}\exp(\Delta F^{i,t})\Delta f_{k}^{i}]\label{exp-min-first}\\
J''|_{\alpha=0} & = & \sum_{i}E_{h|v^{i},x^{i},t}[\sum_{v}\exp(\Delta F^{i,t})(\Delta f_{k}^{i})^{2}]\label{exp-min-second}
\end{eqnarray}
Expanding (\ref{exp-min-first}) yields 
\begin{eqnarray}
J'|_{\alpha=0}=\sum_{i}\frac{1}{p(v^{i}|x^{i},t)}\sum_{v,h}p(v,h|x^{i},t)\Delta f_{k}^{i}\label{exp-min-first2}
\end{eqnarray}
Note that $f_{k}^{i}$ has the additive form as in (\ref{additive-form})
so $\Delta f_{k}(x^{i},y)=\sum_{c}\Delta f_{k}(x^{i},y_{c})$. Thus
(\ref{exp-min-first2}) reduces to 
\begin{eqnarray}
J'|_{\alpha=0}=\sum_{i}\frac{1}{p(v^{i}|x^{i},t)}\sum_{c}\sum_{y_{c}}p(y_{c}|x^{i},t)\Delta f_{k}(x^{i},y_{c})\label{exp-min-first3}
\end{eqnarray}
which now contains clique marginals and can be estimated efficiently
for tree-like structure using a downward and upward sweep. For general
structures, loopy belief propagation can provide approximate estimates.
Details of the procedure are omitted here due to space constraints.

However, the computation of (\ref{exp-min-second}) does not enjoy
the same efficiency because the square function is not decomposable.
To make it decomposable, we employ Cauchy's inequality to yield the
upper bound of the change in (\ref{Taylor-approx}) 
\begin{eqnarray*}
(\Delta f_{k}(x^{i},y))^{2}=(\sum_{c}\Delta f_{k}(x^{i},y_{c}))^{2}\le|C|\sum_{c}\Delta f_{k}(x^{i},y_{c})^{2}
\end{eqnarray*}
where $|C|$ is the number of cliques in the network.

The update using $\alpha=-J'/\tilde{J}''$, where $\tilde{J}''$ is
the upper bound of the second derivative $J''$, is rather conservative,
so it is clear that a further line search is needed. Moreover, it
should be noted that the change in (\ref{Taylor-approx}), due to
the Newton update, is $\Delta\tilde{J}(\alpha,k)=-0.5(J')^{2}/\tilde{J}''$,
where $\Delta\tilde{J}$ is the upper bound of the change $\Delta J$
due to Cauchy's inequality, so the weak learner selection using the
optimal change does not depend on the scale of the second derivative
bound of $\tilde{J}''$. Thus the term $|C|$ in Cauchy's inequality
above can be replaced by any convenient constant.

The complexity of our boosting algorithm is the same as that in the
MLE of the Markov networks. This can be verified easily by taking
the derivative of the log-loss in (\ref{log-loss}) and comparing
it with the quantities required in our algorithm.

%----------

\section{Experimental results}

We restrict our attention to the linear-chain structure for efficient
computation because it is sufficient to learn from the video data
we capture. %All the computation described
%in the previous section can be carried out using a
%single forward-backward pass. 
For all the experiments reported here, we train the model using the
MLE along with the limited memory quasi-Newton method (L-BFGS) and
we use the proposed boosting scheme with the help of a line search,
satisfying Amijo's conditions. For regularisation, the same $\sigma$
is used for all features for simplicity and is empirically selected.
In the training data, only 50\% of labels are randomly given for each
data slice in the sequence. For the performance measure, we report
the per-label error and the average $F1$-score over all distinct
labels.

\subsection{Data and feature extraction}

%-------
\begin{figure}[htb]
\begin{centering}
\begin{tabular}{cc}
\includegraphics[width=0.4\linewidth]{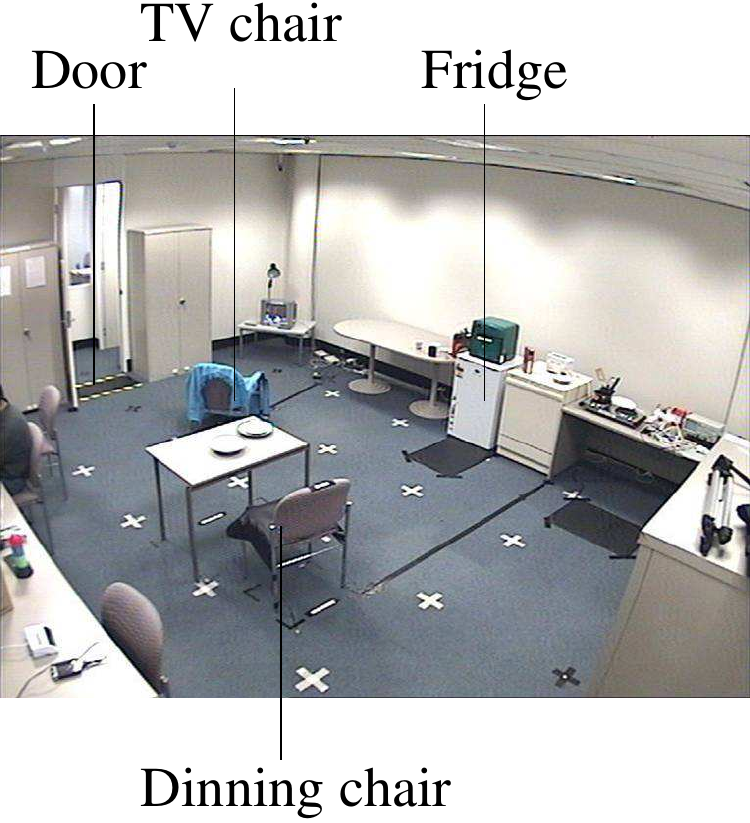} & \includegraphics[width=0.4\linewidth]{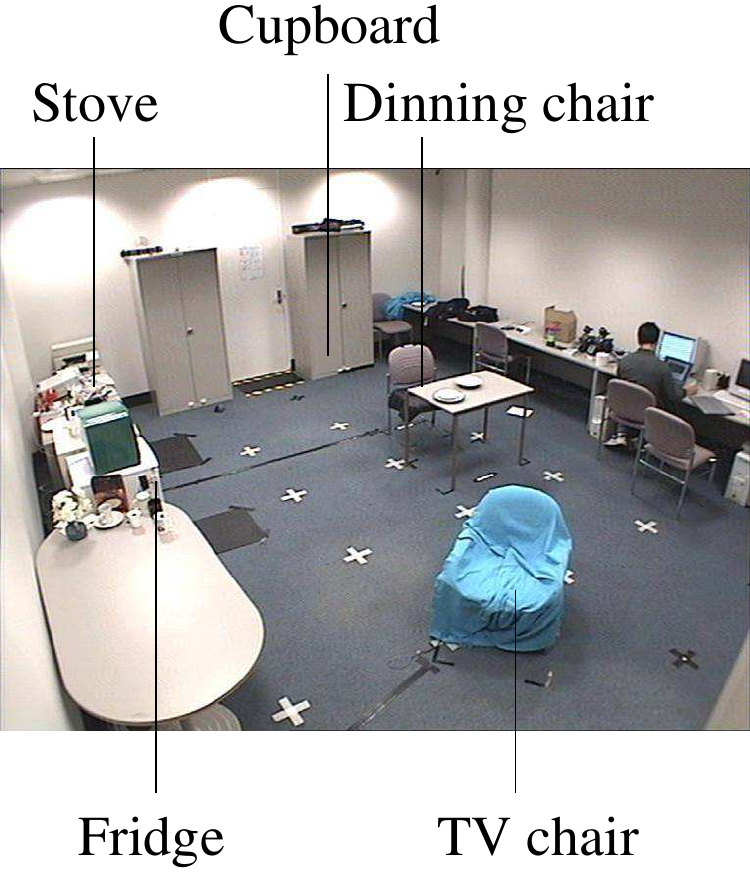}\tabularnewline
\end{tabular}
\par\end{centering}

\caption{The environment and scene viewed from the two cameras.}

\label{fig:cam} 
\end{figure}

In this paper, we evaluate our boosting framework on video sensor
data. However, the framework is applicable to different type of sensors
and is able to fuse different sensor information. The observational
environment is a kitchen and dining room with two cameras mounted
to two opposite ceiling corners (Figure \ref{fig:cam}). The observations
are sequences of noisy coordinates of a person walking in the scene
extracted from video using a background subtraction algorithm. The
data was collected in our recent work \cite{Nam-et-alCVPR05}, consisting
of 90 video sequences spanning three scenarios: SHORT\_MEAL, HAVE\_SNACK
and NORMAL\_MEAL. We pick a slightly smaller number of sequences,
of which we select 42 sequences for training and another 44 sequences
for testing. Unlike \cite{Nam-et-alCVPR05}, we only build flat models,
thus separate models for each scenario are learned and tested. The
labels are sequences of the `activities' the person is performing,
for example \textit{`going-from-door-to-fridge'}. For training, only
partial label sets are given, leaving a portion of the label missing.
For testing, the labels are not provided, and the labels obtained
from the decoding task are compared against the ground-truth.

It turns out that informative features are critical to the success
of the models. At the same time, the features should be as simple
and intuitive as possible to reduce manual labour. At each time slice
$\tau$, we extract a vector of 5 elements from the observation sequence
$g(x,\tau)=(X,Y,u_{X},u_{Y},s=\sqrt{u_{X}^{2}+u_{Y}^{2}})$, which
correspond to the $(X,Y)$ coordinates, the $X$ \& $Y$ velocities,
and the speed, respectively. These observation features are approximately
normalised so that they are of comparable scale.

\subsection{Effect of feature selection}

\label{sec:feature-select} Following previous boosting applications
(e.g.\cite{schapire00boostexter}), we employ very simple decision
rules: a weak hypothesis (feature function) returns a real value if
certain conditions on the training data points are met, and 0 otherwise.

We design three feature sets. The first set, called the \textit{activity-persistence},
captures the fact that activities are in general persistent. The set
is divided into data-association features 
\begin{eqnarray}
f_{l,m}(x,y_{\tau})=\delta[y_{\tau}=l]g_{m}(x,\tau)\label{data-ass-feature}
\end{eqnarray}
where $m=1,..,5$, and label-label features 
\begin{eqnarray}
f_{l,m}(x,y_{\tau-1},y_{\tau})=\delta[y_{\tau-1}=y_{\tau}]\delta[y_{\tau}=l]
\end{eqnarray}
Thus the set has $K=5|Y|+|Y|$ features, where $|Y|$ is the size
of the label set.

The second feature set consists of \textit{transition-features} that
are intended to encode the activity transition nature as follows 
\begin{eqnarray}
f_{l1,l2,m}(x,y_{\tau-1},y_{\tau})=\delta[y_{\tau-1}=l1]\delta[y_{\tau}=l2]g_{m}(x,\tau)
\end{eqnarray}
Thus the size of the feature set is $K=5|Y|^{2}$.

The third set, called the \textit{context set}, is a generalisation
of the second set. Observation-features now incorporate neighbouring
observation points within a sliding window of width $W$ 
\begin{eqnarray}
g_{m}(x,\tau,\epsilon)=g_{m}(x,\tau+\epsilon)
\end{eqnarray}
where $\epsilon=-W_{l},..0,..W_{u}$ with $W_{l}+W_{u}+1=W$. This
is intended to capture the correlation of the current activity with
the past and the future, or the temporal \textit{context} of the observations.
The second feature set is a special case with $W=1$. The number of
features is a multiple of that of the second set, which is $K=5W|Y|^{2}$.

The boosting studied in this section has the beam size $S=1$, i.e.
each round picks only one feature to update its weight. Tables \ref{feature_set1},
\ref{feature_set2}, \ref{feature_set3} show the performance of the
training algorithms on test data of all three scenarios (SHORT\_MEAL,
HAVE\_SNACK and NORMAL\_MEAL) for the three feature sets, respectively.
Note that the infinite regularisation factor $\sigma$ means that
there is no regularisation. In general, sequential boosting appears
to be slower than the MLE because it updates only one parameter at
a time. For the activity persistence features (Table \ref{feature_set1}),
the feature set is very compact but informative enough so that the
MLE attains a reasonably high performance. Due to this compactness,
the feature selection capacity is almost eliminated, leading to poorer
results as compared with the MLE.

However, the situation changes radically for the activity transition
feature set (Table \ref{feature_set2}) and for the context feature
set (Table \ref{feature_set3}). When the observation context is small,
i.e. $W=1$, boosting consistently outperforms the MLE whilst maintaining
only a partial subset of features ($<50\%$ of the original feature
set). The feature selection capacity is demonstrated more clearly
with the context-based feature set ($W$ = 11), where less than 9\%
of features are selected by boosting for the SHORT\_MEAL scenario,
and less than 3\% for the NORMAL\_MEAL scenario. The boosting performance
is still reasonable despite the fact that a very compact feature set
is used. There is therefore a clear computational advantage when the
learned model is used for classification.

%--------
\begin{table}
\caption{Performance on three data sets, activity-persistence features. SM
= SHORT\_MEAL, HS = HAVE\_SNACK , NM = NORMAL\_MEAL, Agthm = algorithm,
itrs = number of iterations, ftrs = number of selected features, \%
ftrs = portion of selected features.}

\begin{centering}
\begin{tabular}{|l||c|c||c|c||c|c|}
\hline 
Data  & SM  & SM  & HS  & HS  & NM  & NM \tabularnewline
\hline 
Agthm  & MLE  & Boost  & MLE  & Boost  & MLE  & Boost \tabularnewline
\hline 
$\sigma$  & $\infty$  & $\infty$ & $\infty$  & $\infty$ & $\infty$  & $\infty$ \tabularnewline
\hline 
error(\%)  & 10.3  & 16.6  & 12.4  & 14.5  & 9.7  & 17.2\tabularnewline
\hline 
F1(\%)  & 86.0  & 80.2  & 84.8  & 82.1  & 87.9  & 77.4 \tabularnewline
\hline 
itrs  & 100  & 500  & 100  & 200  & 100  & 200 \tabularnewline
\hline 
\# ftrs  & 30  & 30  & 30  & 30  & 42  & 35 \tabularnewline
\hline 
\% ftrs  & 100  & 100  & 100  & 100  & 100  & 83.3 \tabularnewline
\hline 
\end{tabular}
\par\end{centering}

\label{feature_set1} 
\end{table}

%--------
\begin{table}
\caption{Performance on activity transition features}

\begin{centering}
\begin{tabular}{|l||c|c||c|c||c|c|}
\hline 
Data  & SM  & SM  & HS  & HS  & NM  & NM \tabularnewline
\hline 
Agthm  & MLE  & Boost  & MLE  & Boost  & MLE  & Boost \tabularnewline
\hline 
$\sigma$  & $\infty$  & $\infty$ & $\infty$  & $\infty$ & $\infty$  & $\infty$ \tabularnewline
\hline 
error(\%)  & 18.6  & 10.1  & 13.0  & 10.8  & 15.0  & 16.5\tabularnewline
\hline 
F1(\%)  & 75.8  & 89.3  & 86.8  & 85.7  & 81.4  & 80.9\tabularnewline
\hline 
itrs  & 59  & 200  & 74  & 100  & 53  & 100\tabularnewline
\hline 
\# ftrs  & 125  & 57  & 125  & 44  & 245  & 60 \tabularnewline
\hline 
\% ftrs  & 100  & 45.6  & 100  & 35.2  & 100  & 24.5 \tabularnewline
\hline 
\end{tabular}
\par\end{centering}

\label{feature_set2} 
\end{table}

%--------
\begin{table}
\caption{Performance on context features with window size $W=11$}

\begin{centering}
\begin{tabular}{|l||c|c||c|c||c|c|}
\hline 
Data  & SM  & SM  & HS  & HS  & NM  & NM \tabularnewline
\hline 
Agthm  & MLE  & Boost  & MLE  & Boost  & MLE  & Boost \tabularnewline
\hline 
$\sigma$  & 2  & 2  & $\infty$  & $\infty$  & $\infty$  & $\infty$ \tabularnewline
\hline 
error(\%)  & 15.3  & 9.6  & 9.4  & 11.2  & 9.3  & 16.6 \tabularnewline
\hline 
F1(\%)  & 81.6  & 87.7  & 89.3  & 86.6  & 87.7  & 78.1\tabularnewline
\hline 
itrs  & 51  & 200  & 22  & 100  & 21  & 100\tabularnewline
\hline 
\# ftrs & 1375  & 115  & 1375  & 84  & 2695  & 80\tabularnewline
\hline 
\% ftrs  & 100  & 8.36  & 100  & 6.1  & 100  & 3.0\tabularnewline
\hline 
\end{tabular}
\par\end{centering}

\label{feature_set3} 
\end{table}

\subsection{Learning the activity-transition model}

We demonstrate in this section that the activity transition model
can be learned by both the MLE and boosting. The transition feature
sets studied previously do not separate the transitions from data,
so the transition model may not be correctly learned. We design another
feature set, which is the bridge between the activity-persistence
and the transition feature set. Similar to the activity persistence
set, the new set is divided into data-association features, as in
(\ref{data-ass-feature}), and label-label features 
\begin{eqnarray}
f_{l1,l2}(y_{\tau-1},y_{\tau})=\delta[y_{\tau-1}=l1]\delta[y_{\tau}=l2]
\end{eqnarray}
Thus the set has $K=5|Y|+|Y|^{2}$ features.

Given the SHORT\_MEAL data set, and the activity transition matrix
in Table \ref{tran-matrix1}, the parameters corresponding to the
label-label features are given in Tables \ref{param-matrix1-boost}
and \ref{param-matrix1-MLE}, as learned by boosting and MLE, respectively.
%--------
\begin{table}
\caption{Activity transition matrix of SHORT\_MEAL data set}

\begin{centering}
\begin{tabular}{|c||c|c|c|c|c|}
\hline 
Activity  & 1  & 2  & 3  & 4  & 11 \tabularnewline
\hline 
\hline 
1  & 1  & 1  & 0  & 0  & 0 \tabularnewline
\hline 
2  & 0  & 1  & 1  & 0  & 1 \tabularnewline
\hline 
3  & 0  & 0  & 1  & 1  & 0 \tabularnewline
\hline 
4  & 0  & 0  & 0  & 1  & 0 \tabularnewline
\hline 
11  & 0  & 0  & 0  & 0  & 1 \tabularnewline
\hline 
\end{tabular}
\par\end{centering}

\label{tran-matrix1} 
\end{table}

%--------

%--------
\begin{table}
\caption{Parameter matrix of SHORT\_MEAL data set learned by boosting}

\begin{centering}
\begin{tabular}{|c||c|c|c|c|c|}
\hline 
Activity  & 1  & 2  & 3  & 4  & 11 \tabularnewline
\hline 
\hline 
1 & 1.8 & 0 & -5904.9 & -5904.9 & 0 \tabularnewline
\hline 
2 & -5904.9 & 3.6 & 0 & -5904.9 & 0 \tabularnewline
\hline 
3 & -5904.9 & -5904.9 & 2.425 & 0 & -5904.9 \tabularnewline
\hline 
4 & -5904.9 & -5904.9 & -5904.9 & 2.4 & -5904.9 \tabularnewline
\hline 
11 & -5904.9 & -5904.9 & -5904.9 & -5904.9 & 2.175 \tabularnewline
\hline 
\end{tabular}
\par\end{centering}

\label{param-matrix1-boost} 
\end{table}

%\begin{table}
%\begin{center}
%\caption{Activity transition matrix of HAVE\_SNACK data set}
%\begin{tabular}[pos=tb]{|c||c|c|c|c|c|}\hline
%Activity &	5&	6&	2&	7&	8 \\ \hline \hline
%5	&	1&	1&	0&	0&	1
%6	&	0&	1&	1&	0&	0	 \\ \hline
%2	&	0&	0&	1&	1&	0 \\ \hline
%7	&	0&	1&	0&	1&	1 \\ \hline
%8	&	0&	0&	0&	0&	1 \\ \hline
%\end{tabular}
%\end{center}
%\label{tran-matrix1}
%\end{table}

%\begin{table}
%\caption{Parameter matrix of HAVE\_SNACK data set learned by boosting}
%\begin{center}
%\begin{tabular}[pos=tb]{|c||c|c|c|c|c|}\hline
%Activity &	5&		6&		2&		7&		8 \\ \hline \hline
%5	&	1.25&		0&		-5904.9&	-5904.9&	0 \\ \hline
%6	&	-5904.9&	2.625&		0&		-5904.9&	-5904.9	 \\ \hline	
%2	&	-5904.9&	-5904.9&	3.525&		0&		-5904.9	 \\ \hline
%7	&	-5904.9&	0&		-5904.9&	1.625&		0	 \\ \hline
%8	&	-5904.9&	-5904.9&	-5904.9&	-5904.9&	-0.15 \\ \hline
%\end{tabular}
%\end{center}
%\label{tran-matrix1}
%\end{table}

At first sight, it may be tempting to select non-zero parameters and
their associated transition features, and hence the corresponding
transition model. However, as transition features are non-negative
(indicator functions), the model actually penalises the probabilities
of any configurations that activate negative parameters exponentially,
since $p(y|x)\propto\exp(\lambda_{k}f_{k}(y_{\tau-1},y_{\tau}))$.
Therefore, huge negative parameters practically correspond to improbable
configurations. If we replace all non-negative parameters in Table
\ref{param-matrix1-boost} and \ref{param-matrix1-MLE} by 1, and
the rest by 0, we actually obtain the transition matrix in Table \ref{tran-matrix1}.
The difference between boosting and MLE is that boosting penalises
the improbable transitions much more severely, thus leading to much
sharper decisions with high confidence. Note that for this data set,
boosting learns a much more correct model than the MLE, with an error
rate of 3.8\% ($F1$ = 93.7\%), in constrast to 15.6\% ($F1$ = 79.5\%)
by the MLE without regularisation, and 11.8\% ($F1$ = 85.0\%) by
the MLE with $\sigma=5$.

\begin{table}
\caption{Parameter matrix of SHORT\_MEAL data set learned by MLE}

\begin{centering}
\begin{tabular}{|c||c|c|c|c|c|}
\hline 
Activity  & 1  & 2  & 3  & 4  & 11 \tabularnewline
\hline 
\hline 
1 & 10.81  & 4.311  & -5.7457  & -5.3469  & -1.8398 \tabularnewline
\hline 
2 & -2.2007  & 15.056  & 3.6388  & -5.6644  & 0.41921 \tabularnewline
\hline 
3 & -5.3565  & -2.3131  & 9.3656  & 1.6575  & -2.3736 \tabularnewline
\hline 
4 & -5.4103  & -4.556  & -4.1142  & 7.1332  & -5.2976 \tabularnewline
\hline 
11 & -3.17  & -0.09001  & -2.9518  & -4.8741  & 8.9128 \tabularnewline
\hline 
\end{tabular}
\par\end{centering}

\label{param-matrix1-MLE} 
\end{table}

\subsection{Effect of beam size}

Recall that the beam search described in Section \ref{sec:beam} allows
the weak hypothesis to be an ensemble of $S$ features. When $S=K$,
all the parameters are updated in parallel, so it is essentially similar
to the MLE, and thus no feature selection is performed. We run a few
experiments with different beam sizes $S$, starting from 1, which
is the main focus of this paper, to the full parameter set $K$. As
$S$ increases, the number of selected features also increases. However,
we are quite inconclusive about the final performance. It seems that
when $S$ is large, the update is quite poor, leading to slow convergence.
This is probably because the diagonal matrix resulting from the algorithm
is not a good approximation to the true Hessian used in Newton's updates.
It suggests that there exists a good, but rather moderate, beam size
that performs best in terms of both the convergence rate and the final
performance.

An alternative is just to minimise the exponential loss in (\ref{exp-loss})
directly by using any generic optimisation method (e.g. see \cite{AA63,Altun-03}).
%This has been tried previously in the context
%of fully observed data in \cite{AA63, Altun-03}. 
%The authors
%reported comparable performance as compared to the MLE.
However, this approach, although may be fast to converge, loses the
main idea behind boosting, which is to re-weigh the data distribution
on each round to focus more on hard-to-classify examples as in (\ref{weighted-loss}).
These issues are left for future investigation.

%\subsection{Training vs test error}
%----------

\section{Conclusions and Further work}

We have presented a scheme to exploit the discriminative learning
power of the boosting methodology and the semantically rich structured
model of Markov networks and integrated them into a boosted Markov
network framework which can handle missing variables. We have demonstrated
the performance of the newly proposed algorithm over the standard
maximum-likelihood framework on video-based activity recognition tasks.
Our preliminary results on structure learning using boosting indicates
promise. Moreover, the built-in capacity of feature selection by boosting
suggests an interesting application area in small footprint devices
with limited processing power and batteries. We plan to investigate
how to select the optimal feature set online by hand-held devices
given the processor, memory and battery status.

Although empirically shown to be successful in our experiments, the
performance guarantee of the framework is yet to be proven, possibly
following the large margin approach as in \cite{Sch98,NIPS2003_AA04},
or the asymptotic consistency in the statistics literature as with
the MLE. In the application to sensor networks, we intend to explore
methods to incorporate richer sensory information into the weak learners,
and to build more expressive structures to model multi-level and hierarchical
activities.

% ACKNOWLEDGMENTS AND REFERENCES

\section*{Acknowledgments}

% note that here \section* is used instead of \section
The Matlab code of the limited memory quasi-Newton method (L-BFGS)
to optimise the log-likelihood of the CRFs is adapted from S. Ulbrich.

\end{document}